\documentclass[letterpaper, 10 pt, conference]{ieeeconf}

\IEEEoverridecommandlockouts             
\usepackage{graphics} 
\usepackage{epsfig} 
\usepackage{times} 
\usepackage{amsmath} 
\usepackage{amssymb}  
\usepackage[per-mode=symbol]{siunitx}
\usepackage{hyperref}
\usepackage{algorithm}
\usepackage{algpseudocode}
\usepackage{listings}
\usepackage[font=footnotesize]{caption}
\usepackage[isbn=false,
    backend=biber,
    style=ieee,
    bibstyle=ieee,  
    sortcites=true,   
    giveninits=true,            
    backref=false]{biblatex}
\usepackage{xcolor}

\definecolor{codegreen}{rgb}{0,0.6,0}
\definecolor{codegray}{rgb}{0.5,0.5,0.5}
\definecolor{codepurple}{rgb}{0.58,0,0.82}
\definecolor{backcolour}{rgb}{0.95,0.95,0.92}

\lstdefinestyle{mystyle}{
  backgroundcolor=\color{backcolour}, commentstyle=\color{codegreen},
  keywordstyle=\color{magenta},
  numberstyle=\tiny\color{codegray},
  stringstyle=\color{codepurple},
  basicstyle=\ttfamily\scriptsize,
  breakatwhitespace=false,         
  breaklines=true,                 
  captionpos=b,                    
  keepspaces=true,                 
  showspaces=false,                
  showstringspaces=false,
  showtabs=false,                  
  tabsize=2
}
\let\Re\relax
\DeclareMathOperator{\Re}{\mathbb{R}}

\DeclareMathOperator{\xvec}{\mathbf{x}}
\DeclareMathOperator{\uvec}{\mathbf{u}}

\DeclareMathOperator{\Xvec}{\mathbf{z}}
\DeclareMathOperator{\Geq}{\mathbf{G}^{eq}}
\DeclareMathOperator{\Gineq}{\mathbf{G}^{ineq}}

\title{\LARGE \bf CusADi: A GPU Parallelization Framework for \\ 
Symbolic Expressions and Optimal Control}

\author{Se Hwan Jeon$^{1}$, Seungwoo Hong$^{1}$, Ho Jae Lee$^{1}$, Charles Khazoom$^{1}$, and Sangbae Kim$^{1}$%
\thanks{$^{1}$Department of Mechanical Engineering, Massachusetts Institute of Technology, Cambridge, MA 02139, USA
{\tt\small \{sehwan, swhong, hjlee201, ckhaz, sangbae\}@mit.edu}}%
}

\AtBeginBibliography{\footnotesize}

\begin{document}

\maketitle
\thispagestyle{empty}
\pagestyle{empty}

\allowdisplaybreaks
\newcommand{\algorithmautorefname}{Algorithm}
\renewcommand{\figureautorefname}{Fig.}
\renewcommand{\equationautorefname}{Eq.}
\renewcommand{\sectionautorefname}{Section}
\renewcommand{\subsectionautorefname}{Section}

\newcommand{\casadi}{\texttt{casadi}}
\newcommand{\cusadi}{\texttt{CusADi}}

\newcommand{\comment}[1]{\textcolor{blue}{#1}}
\newcommand{\hojae}[1]{{\color{cyan}Ho Jae: #1}}

\begin{abstract}
The parallelism afforded by GPUs presents significant advantages in training controllers through reinforcement learning (RL).
However, integrating model-based optimization into this process remains challenging due to the complexity of formulating and solving optimization problems across thousands of instances.
In this work, we present \cusadi, an extension of the \casadi\ symbolic framework to support the parallelization of arbitrary closed-form expressions on GPUs with \texttt{CUDA}.
We also formulate a closed-form approximation for solving general optimal control problems, enabling large-scale parallelization and evaluation of MPC controllers.
Our results show a ten-fold speedup relative to similar MPC implementation on the CPU, and we demonstrate the use of \cusadi\ for various applications, including parallel simulation, parameter sweeps, and policy training.
\end{abstract}

\section{INTRODUCTION}

Using GPUs for robotics is attractive due to their powerful computing and parallelization capabilities compared to CPUs.
These advantages are particularly beneficial in training controllers through parallelized simulations and reinforcement learning (RL), evidenced by the success of learned policies in handling complex, high-dimensional tasks \cite{Miki2022_LearningLocomotion,Cheng2024_ExtremeParkourLegged,Zhuang2023_RobotParkour,Hoeller2020_DeepValueMPC}.
With cheaper compute, it is appealing to begin incorporating model-based techniques and optimization into training, where the sample efficiency, exploration, and interpretability of the policy could all be improved by embedding model-based domain knowledge as part of the learning pipeline \cite{Jenelten2024_DTC, Grandesso2023_CACTO,Lee2024_RLHumanoidLIPPlanning}.

Moreover, the barrier to creating model-based controllers has been substantially lowered.
There exists an ecosystem of software tools that simplify developing, designing, and tuning controllers, such as \texttt{OCS2}, \texttt{Crocoddyl}, \texttt{rockit}, and \casadi\cite{OCS2, Mastalli2020_crocoddyl, Gillis2020_rockit, Andersson2019_casadi}.
\casadi's symbolic framework in particular greatly simplifies the process of formulating the costs, constraints, and dynamics of an optimal control problem (OCP).

However, it is difficult to embed these controllers directly into learning environments because these tools are confined to CPU evaluation.
Solving optimization problems across thousands of RL agents is complex to implement on the GPU, and computing their solutions efficiently is more challenging still.
Generally, GPU parallelization has been used to speedup a single "large" numerical problem by exploiting repeated structures within it (e.g., long-horizon model predictive control (MPC) or a system with high-dimensional states).
These can often be decomposed into independent, parallelizable subproblems, such as computing gradients of constraints in trajectory optimization \cite{Hyatt2017_GPUevolutionaryMPC,Plancher2021_DynamicsGPU,Plancher2019_DDPGPU,Plancher2020_ParallelDDPGPU} or the matrix factorizations for solving linear systems \cite{Adabag2023_MPCGPU,Schubiger2020_cuOSQP, Kang2024_FastCertifiableTO}.
These works are specialized to parallelize specific aspects of their numerical problem.

For RL applications however, the parallelization needed for computational efficiency is not \textit{within} a single instance or controller, but rather \textit{across} the thousands of environments in simulation.
There are relatively few works that extend computational tools for batch evaluation on the GPU.
\textcite{Amos2017_OptNet} developed a custom solver for batches of small QPs on the GPU, but does not exploit sparsity patterns present in MPC, and is specialized for solving small dense linear systems in batches.
Frameworks like \texttt{PyTorch} and \texttt{JAX} similarly lack mature libraries for sparse matrix algebra \cite{Bradbury2018_JAX,Paszke2019_Pytorch}.
Although these computations could also be offloaded to the CPU, the limited number of threads and the overhead incurred by CPU-GPU data transfer make this inefficient.
For example, solving MPC on the CPU in RL training loops can take weeks for full policy convergence \cite{Jenelten2024_DTC}.

\begin{figure}[t]
    \centering
    \includegraphics[width=1\linewidth]{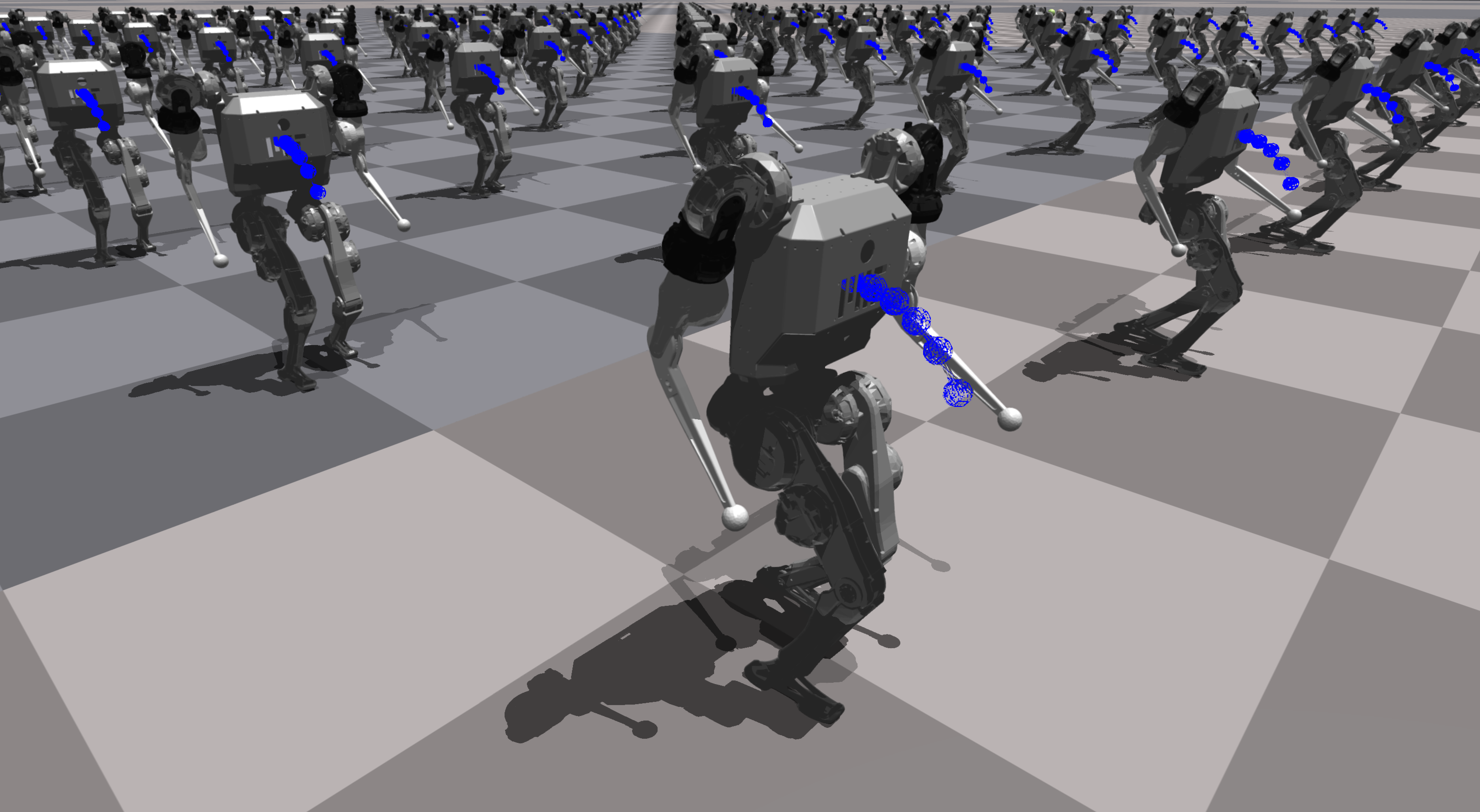}
    \caption[]{Parallelizing MPC for the MIT Humanoid across thousands of environments in NVIDIA's IsaacGym \cite{Makoviychuk2021_isaacgym}. The predicted positions of the base are shown as blue spheres.}
    \label{fig:parallel_MPC}
    \vspace{-6mm}
\end{figure}

In this work, we present \cusadi, an extension of the \casadi\ symbolic framework with \texttt{CUDA} for parallel evaluation on the GPU\footnote{\href{https://github.com/se-hwan/cusadi}{Repository and videos}: https://github.com/se-hwan/cusadi}
\cusadi\ code-generates and compiles symbolic functions from \casadi, enabling parallel evaluation for any specified batch size.
Algorithms and optimizations formulated symbolically for a single instance can then be evaluated simultaneously for thousands on the GPU.
\cusadi\ serves as a bridge for embedding model-based techniques and expressions from \casadi\ into RL environments, offering speedups of up to 10-100x compared to parallel CPU evaluation, depending on data transfer overhead.

We show several examples highlighting robotics applications with \cusadi.
First, we formulate a closed-form approximation to the OCP that is amenable for parallelization and deploy MPC across thousands of environments in IsaacGym \cite{Makoviychuk2021_isaacgym}, as shown in \autoref{fig:parallel_MPC}, with training iterations roughly 11x faster than in \cite{Jenelten2024_DTC}.
Second, we demonstrate how dynamic quantities, such as the centroidal momentum or composite rigid-body inertia, can be symbolically expressed in \casadi, computed in parallel with \cusadi, and used to augment the observations and rewards in a training environment.
Lastly, we run custom parallel simulations for a planar quadcopter system to efficiently evaluate parameter sensitivity and the region of attraction.

We summarize our main contributions as follows:
\begin{enumerate}
    \item We present \cusadi, our open-source tool to parallelize arbitrary symbolic functions on the GPU.
    \item We formulate a closed-form approximation to the optimal control problem for GPU parallelization.
    \item We demonstrate how \cusadi\ can be used for various robotics applications, including parallelized simulation, parameter sweeps, and reinforcement learning.
\end{enumerate}

\section{Background}

\begin{figure*}[tb]
    \centering
    \includegraphics[width=0.9\textwidth]{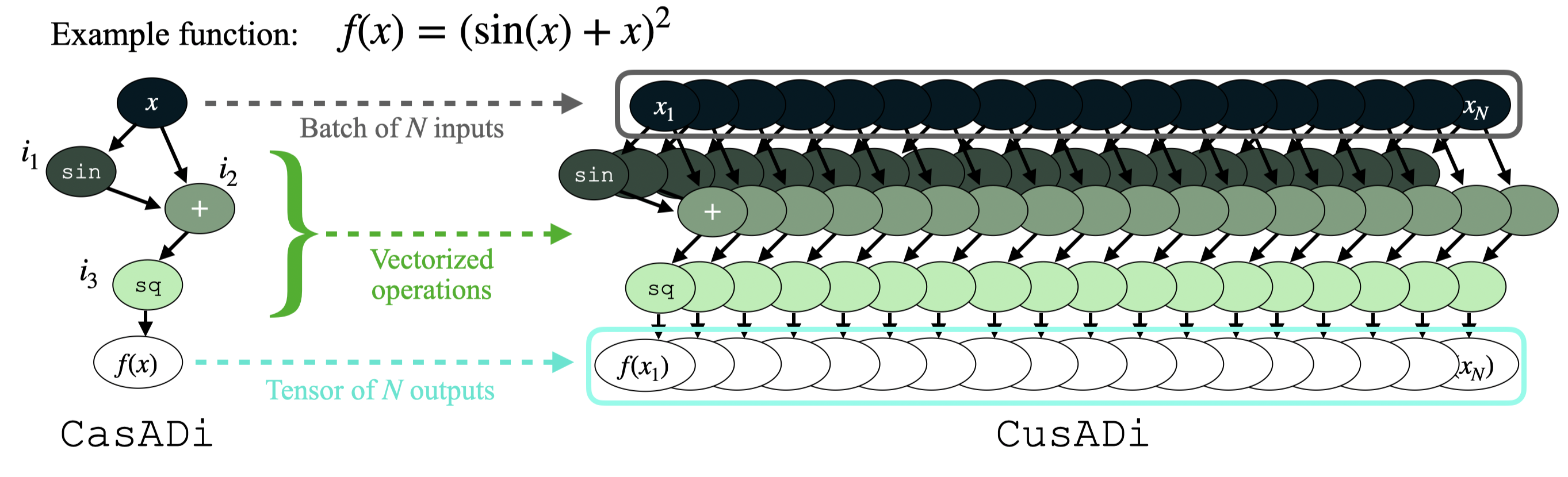}
    \caption{Visualization of \cusadi\ parallelization. \textit{Left:} Symbolic expressions in \casadi\ are represented as expression graphs, a sequence of atomic operations ($i_1, i_2, i_3$) which evaluate the function.
    \textit{Right:} Each atomic operation in the sequence can be vectorized to act on an arbitrary number of elements with \texttt{CUDA}; by repeating this for all operations in the original expression, \casadi\ symbolic expressions can be evaluated for thousands of instances in parallel on the GPU.}
    \label{fig:cusadi_graph}
\end{figure*}

\subsection{\casadi}
\label{section:casadi}

\casadi\ is a software stack designed for gradient-based numerical optimization and is widely used for optimal control \cite{Andersson2019_casadi}.
Symbolic expressions in \casadi\ are one of two data types: \texttt{SX} or \texttt{MX}.
\texttt{SX} expressions in \casadi\ are represented as directed graphs where each node represents an \textit{atomic operation}, as shown in \autoref{fig:cusadi_graph} (left).
These atomic operations are either unary ($\Re \rightarrow \Re$) or binary ($\Re \times \Re \rightarrow \Re$), and arbitrary closed-form expressions can be expressed as a finite sequence of these scalar operations.
Examples of unary operations are \texttt{log}, \texttt{cos}, and \texttt{sqrt}, and examples of binary operations are addition, multiplication, and \texttt{atan2}.
Note that this does not limit \texttt{SX} expressions to scalar inputs or outputs; operations such as matrix multiplication are simply expanded into unary and binary ones between scalar elements of the function.

The \texttt{MX} type generalizes the \texttt{SX} type and consists of sequences of operations that are not limited to be scalar unary or binary operations. 
Matrix expressions can also be transformed into a series of atomic operations on scalar values with the \texttt{expand} functionality in \casadi.

We chose the \casadi\ stack for parallelization for several reasons.
Firstly, \casadi\ fully supports sparse matrix algebra and algorithmic differentiation (AD), ensuring expressions are both efficient and differentiable.
This makes it convenient to take Jacobians and Hessians symbolically.
While other frameworks such as \texttt{Pytorch} and \texttt{JAX} also support symbolic and differentiable functions, sparse operations are not yet fully mature in either \cite{Paszke2019_Pytorch, Bradbury2018_JAX}.
Exploiting sparsity is crucial given the structure present in optimal control problems, where typically only a small fraction of the Karush-Kuhn-Tucker (KKT) system has non-zero elements \cite{Betts1999_SparsityInOCP}.
With \casadi, the expression graphs compute only the non-zero outputs for symbolic functions.

Secondly, the \texttt{Opti} stack in \casadi\ streamlines the process of defining the variables and parameters of an optimal control problem, providing convenient interfaces to a breadth of solvers, including 
\texttt{IPOPT}, \texttt{KNITRO}, and \texttt{OSQP} \cite{Wachter2002_IPOPT,Byrd2006_Knitro,Stellato2020_OSQP}.

Lastly, several robotic toolboxes available are already compatible with \casadi, such as \texttt{spatial\_v2} \cite{Featherstone2007_RBDA}, \texttt{Pinocchio} \cite{Carpentier2019_Pinocchio}, and \texttt{GRBDA} \cite{Chignoli2023_GRBDA}.
Rewriting these dynamic libraries and algorithms in a different symbolic framework would require significant and largely unnecessary effort.
Expressions computed from these libraries can be exported as a \casadi\ expression graph directly callable from \texttt{MATLAB}, Python, or \texttt{C++}.

\subsection{GPU Parallelization}
While CPUs have dozens of cores intended for high-speed sequential processing and computation, GPUs consist of thousands of smaller cores with simpler control logic.
Consequently, GPUs excel at performing identical operations on large volumes of data.
This so-called "single instruction multiple data" (SIMD) parallelism allows GPUs to process many data elements simultaneously, dramatically increasing throughput for parallelizable tasks.
Naturally, this architecture has been particularly advantageous in applications large-scale, repetitive computations such as reinforcement learning, graphics processing, and numerical simulation \cite{Patterson_ComputerArchitecture}.
With interfaces such as NVIDIA's \texttt{CUDA} library, users can directly write programs (kernels) to be evaluated in parallel across the threads of a GPU \cite{NVIDIA_CUDA}.

In \autoref{section:cusadi}, we detail how \cusadi\ code-generates \texttt{CUDA} kernels from symbolic \casadi\ expressions.
These kernels are compiled as an externally callable \texttt{C} library with an interface to \texttt{PyTorch}.

\subsection{Optimal Control}
\label{sec:SQP}
We express the discrete, finite-time horizon optimal control problem (OCP) as
\begin{eqnarray}
\label{eq:OCP}
    \min_{\xvec[\cdot], \uvec[\cdot]}   && l_T(\xvec_T, \uvec_T) + \sum_{i=0}^{T-1} l_i(\xvec_i, \uvec_i) \\
    \text{s.t.}             && \xvec_0 = \bar{\xvec}_0\\
                            && \xvec_{i+1} = \mathbf{f}(\xvec_i, \uvec_i), \quad i=0,\dots, T-1 \nonumber  \\
                            && \mathbf{g}^{eq}_i(\xvec_i, \uvec_i) = 0, \quad i=0, \dots, T \nonumber \\
                            && \mathbf{g}^{ineq}_i(\xvec_i, \uvec_i) \leq 0, \quad i=0, \dots , T\nonumber
\end{eqnarray}
where $\xvec[\cdot] \in \Re^{n_x \times T}$ and $\uvec[\cdot] \in \Re^{n_u \times T}$ are the state and control trajectories with initial condition $\bar{\xvec}_0$ and $\mathbf{f}: \Re^{n_x} \times \Re^{n_u} \rightarrow \Re^{n_x}$ is the dynamics of the system.
The system is constrained by equality constraints $\mathbf{g}^{eq}_i:\Re^{n_x} \times \Re^{n_u} \rightarrow \Re^{m_{eq}}$ and inequality constraints $\mathbf{g}^{ineq}_i:\Re^{n_x} \times \Re^{n_u} \rightarrow \Re^{m_{ineq}}$ at timestep $i$.

The problem minimizes the sum of the stage costs $l_i: \Re^{n_x} \times \Re^{n_u}\rightarrow \Re$ and terminal cost $l_T: \Re^{n_x} \times \Re^{n_u}\rightarrow \Re$ over the full horizon $T$.
By defining $\Xvec$, $\Geq$, $\Gineq$, and $J$ as 
\begin{eqnarray}
    &&\Xvec := (\xvec_0, \dots, \xvec_T, \uvec_0, \dots, \uvec_T)             \in \Re^{N}, \\
    &&\Geq : \Re^{N}\rightarrow \Re^{M_{eq}} :=
             (\mathbf{g}^{eq}_0, \dots, \mathbf{g}^{eq}_T), \\
    &&\Gineq : \Re^{N}\rightarrow \Re^{M_{ineq}} :=
               (\mathbf{g}^{ineq}_0, \dots, \mathbf{g}^{ineq}_T),\\
    && J: \Re^{N}\rightarrow \Re := l_T(\xvec_T, \uvec_T) +
          \sum_{i=0}^{T-1} l_i(\xvec_i, \uvec_i),
\end{eqnarray}
the problem can be written as a standard NLP of the form:
\begin{eqnarray}
\label{eq:OCP_general}
    \min_{\Xvec}            && J(\Xvec) \\
    \text{s.t.}             && \Geq(\Xvec) = 0 \nonumber  \\
                            && \Gineq(\Xvec) \leq 0. \nonumber 
\end{eqnarray}

\subsection{Sequential Quadratic Programming}

With some initial guess of $\Xvec_0$, the equality-constraint Lagrange multipliers $\boldsymbol{\lambda}_0$, and the inequality-constraint Lagrange multipliers $\boldsymbol{\sigma}_0$, the sequential quadratic programming (SQP) approach solves a sequence of quadratic programs (QPs) to converge to a locally optimal solution.
The subproblems are formed by taking quadratic models of the objective and linearizations of the constraints about the current guess $\mathbf{v}_k := (\Xvec_k, \boldsymbol{\lambda}_k, \boldsymbol{\sigma}_k)$.
Defining the step direction from the current guess as $\delta\mathbf{v}$, the subproblem takes the form
\begin{eqnarray}
\label{eq:SQP_subproblem}
    \min_{\delta\mathbf{z}} &&  \frac{1}{2} \delta\mathbf{z}^T \mathbf{P}_k \delta\mathbf{z}
                                + \mathbf{c}_k^T \delta\mathbf{z} \\
    \text{s.t.}             &&  \mathbf{A}_k^{eq} \delta\mathbf{z} = \mathbf{b}_k^{eq}, \nonumber \\
                            &&  \mathbf{A}_k^{ineq} \delta\mathbf{z} \leq \mathbf{b}_k^{ineq}, \nonumber
\end{eqnarray}
where $\mathbf{P}_k = \nabla_{\mathbf{z}\mathbf{z}}^2 \mathcal{L}(\mathbf{v}_k)$ and $\mathbf{c}_k = \nabla_{\mathbf{z}} \mathcal{L}(\mathbf{v}_k)$ are the Hessian and gradient of the Lagrangian function, respectively; $\mathbf{A}_k^{eq} = \nabla_{\mathbf{z}}\Geq(\Xvec_k)^T$, $\mathbf{b}_k^{eq}=-\Geq(\Xvec_k)$, $\mathbf{A}_k^{ineq} = \nabla_{\mathbf{z}}\Gineq(\Xvec_k)^T$, and $\mathbf{b}_k^{ineq}=-\Gineq(\Xvec_k)$.

After solving the KKT equations of the QP problem in~\eqref{eq:SQP_subproblem} to obtain step direction $\delta\mathbf{v}_k = (\delta\Xvec_k, \delta\boldsymbol{\lambda}_k, \delta\boldsymbol{\sigma}_k)$, the solution is updated as $\mathbf{v}_{k+1} = \mathbf{v}_k + \alpha\delta\mathbf{v}_k$, where $\alpha$ is a scalar that determines the acceptable step length via backtracking line search methods, such as the Armijo method \cite{Armijo1966_LineSearch}.
The matrices of the QP subproblem are recomputed with the updated solution and \eqref{eq:SQP_subproblem} is resolved.
This process is repeated until the solution and/or cost converges.

In \autoref{sec:application_MPC}, we present an approximate SQP algorithm that can be expressed in closed-form for GPU parallelization.

\section{\cusadi}
\label{section:cusadi}
Our work, which we call \cusadi, leverages \texttt{CUDA} kernels and the graph structure of \casadi\ functions to compute \textit{any} symbolic expression from \casadi\ in parallel with \texttt{CUDA}.
The key insight is that the sequence of atomic operations that define a function can be vectorized to operate on \textit{tensors} of data instead of individual scalar values, as shown in \autoref{fig:cusadi_graph} (right).
By writing each vectorized atomic operation sequentially as a \texttt{CUDA} kernel, thousands of function instances can be computed in parallel, limited only by the memory capacity of the GPU and compilation time.
Unlike prior works, we also assume all incoming and outgoing data are stored only on the GPU so that no additional time is spent checking or transferring data between devices \cite{Schubiger2020_cuOSQP, Jenelten2024_DTC}.

\subsection{Code Generation}
A symbolic \casadi\ function consists of a \texttt{work} vector to store intermediate values, and \texttt{n\_instructions}, where each instruction contains three elements:
\begin{itemize}
    \item \texttt{instruction\_id} (the operation type)
    \item \texttt{instruction\_input} (the operation input index)
    \item \texttt{instruction\_output} (the operation output index)
\end{itemize}
The function is evaluated by traversing across the instructions sequentially and performing each operation on the specified indices.\footnote{\href{https://github.com/casadi/casadi/blob/main/docs/examples/python/accessing_sx_algorithm.py}{Python example} available at: https://github.com/casadi/casadi/}

Instead of traversing the instructions for evaluation, we programmatically generate strings of \texttt{CUDA} code at each iteration.
To do so, we create a map between the \casadi\ operation IDs and their equivalent, vectorized counterparts written in \texttt{CUDA} with the appropriate indices, as shown in \autoref{lst:cusadi_dict}.

\begin{lstlisting}[
    float,
    language=Python,
    caption={Subset of the mappings from \texttt{CasADi} instructions keys (\texttt{instruction\_id}) to strings of \texttt{CUDA} kernels that vectorize the corresponding operation. \texttt{env\_idx} corresponds to the thread index of the kernel, added to the appropriate input/output (\texttt{\%d}) index of the \texttt{work} vector.
    % \vspace{-6mm}
    },
    label=lst:cusadi_dict
]
OP_CUDA_DICT = {
  OP_ASSIGN: "work[env_idx + %d] = work[env_idx + %d];",
  OP_ADD: "work[env_idx + %d] = work[env_idx + %d] + work[env_idx + %d];",
  OP_SUB: "work[env_idx + %d] = work[env_idx + %d] - work[env_idx + %d];",
  OP_MUL: "work[env_idx + %d] = work[env_idx + %d] * work[env_idx + %d];",
  OP_DIV: "work[env_idx + %d] = work[env_idx + %d] / work[env_idx + %d];",
  OP_NEG: "work[env_idx + %d] = -work[env_idx + %d];",
  OP_EXP: "work[env_idx + %d] = exp(work[env_idx + %d]);",
  OP_LOG: "work[env_idx + %d] = log(work[env_idx + %d]);",
  OP_POW: "work[env_idx + %d] = pow(work[env_idx + %d], work[env_idx + %d]);",
  ...
\end{lstlisting}
\begin{lstlisting}[
    float,
    language=C,
    caption={Automatically generated \texttt{CUDA} code for the example \texttt{CasADi} function in \autoref{fig:cusadi_graph}. A unique thread \texttt{idx} is assigned for processing data in parallel, calculated from a local thread index and global block index coordinates. The \texttt{if} statement ensures that the instructions do not operate on data outside of the allocated \texttt{batch\_size}.
    % \vspace{-6mm}
    },
    label=lst:cuda
]
__global__ void evaluate_kernel (
        const double *inputs[],
        double *work,
        double *outputs[],
        const int batch_size) {

    int idx = blockIdx.x * blockDim.x + threadIdx.x;
    int env_idx = idx * n_w;
    if (idx < batch_size) {
        work[env_idx + 0] = inputs[0][idx * nnz_in[0] + 0];
        work[env_idx + 1] = sin(work[env_idx + 0]);
        work[env_idx + 1] = work[env_idx + 1] + work[env_idx + 0];
        work[env_idx + 1] = work[env_idx + 1] * work[env_idx + 1];
        outputs[0][idx * nnz_out[0] + 0] = work[env_idx + 1];
    }
}
\end{lstlisting}

After iterating through and vectorizing each instruction, the code is output to a file and compiled with the \texttt{CUDA Toolkit} in \texttt{C} \cite{NVIDIA_CUDA}.
The programmatically generated code file is shown in \autoref{lst:cuda}.
This process only needs to be done once, offline, and the compiled library can be called for evaluation in any \texttt{CUDA} compatible environment.
By including all the operations in a single kernel, the overhead of starting threads is minimized.

This vectorized unrolling of the operations could be written in higher-level languages, but directly compiling low-level \texttt{CUDA} kernels offers the fastest evaluation without any interpreter overhead.

\subsection{PyTorch Interface}
To access the generated \texttt{CUDA} kernels conveniently, we use \texttt{PyTorch} to call the compiled libraries and store the input/output of vectorized expressions.
The software has a mature library of tensor operations that make it convenient to allocate tensors and interface with data on the GPU.
Furthermore, this allows \cusadi\ to easily be integrated into RL environments such as \cite{Makoviychuk2021_isaacgym}, as demonstrated in \autoref{sec:applications}.
Usage examples and tutorials are available in the \cusadi\ repository.

\begin{figure*}[tb]
    \centering
    \includegraphics[width=1\textwidth]{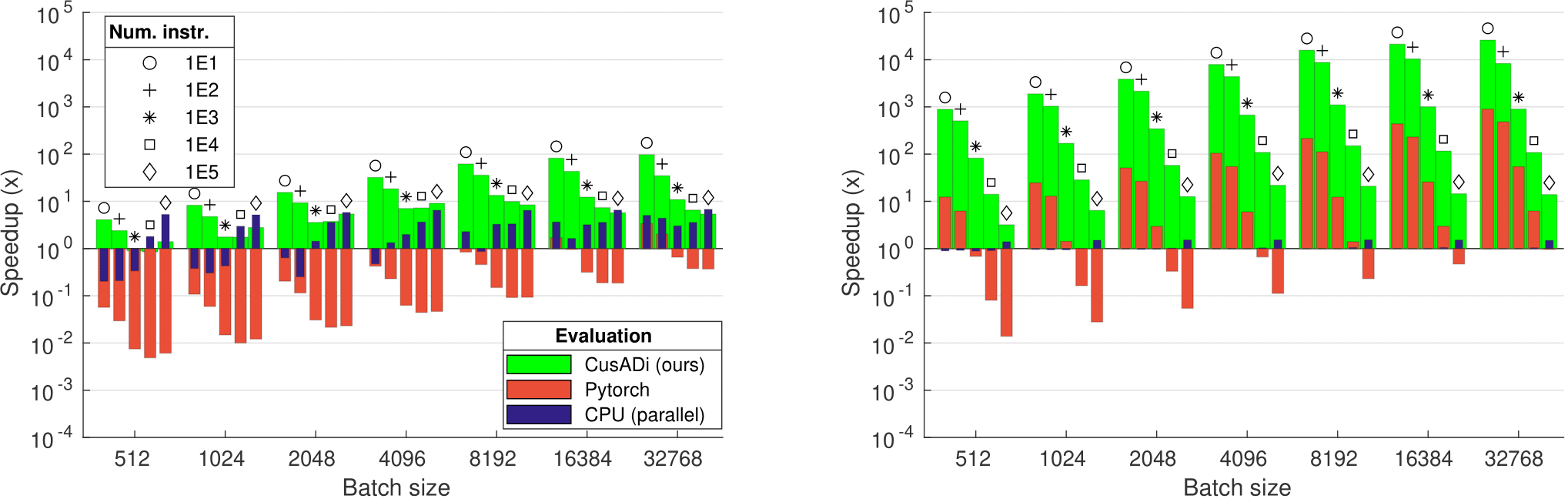}
    \caption{\textit{Left:} Relative speed compared to serial CPU evaluation. The complexity of the function significantly affects the potential speedups from the GPU. \textit{Right:} Relative speed compared to serial CPU evaluation with data transfer overhead. Copying memory between host and memory devices has a substantial effect on speed that is emphasized at larger batch sizes.}
    \label{fig:benchmark}
\end{figure*}
\section{Benchmarking}
\label{section:benchmark}
To evaluate the speedups offered by our GPU parallelization framework, we benchmark the wall clock time of \cusadi\ against serial CPU evaluation, parallel CPU evaluation with the multiprocessing library \texttt{OpenMP}, and with \texttt{PyTorch}.
In the same manner as \autoref{lst:cuda}, we code generate the \casadi\ function in \texttt{PyTorch} for a single instance and evaluate it as a batch with \texttt{PyTorch}'s \texttt{vmap} vectorization method.\footnote{We also test with the new \texttt{torch.compile} functionality introduced in \texttt{PyTorch 2.0}. While the speedups are comparable with those of \cusadi, the initial JIT compilation of the function can require hours to process, and system recursion limits were hit for functions with more than 1,000 operations.}
All benchmarks and applications in \autoref{sec:applications} are conducted on a desktop computer with an Intel i7 10850K processor and NVIDIA 3090 GPU.

We compare five functions, each with an order of magnitude more operations than the previous one, across a range of batch sizes.
Each function evaluates the \textit{LDL$^{T}$} decomposition solution to a positive definite linear system \cite{Boyd2004_ConvexOptimization}.
The speedup of \cusadi\ with respect to serial CPU evaluation (with and without data transfer overhead) is shown in \autoref{fig:benchmark}.

The speedups depend on the complexity of the function (i.e., the total number of operations required) and the batch size.
As the batch size increases, the GPU can take advantage of its threads, and parallelization enables speedups that are 1000x faster than serial CPU evaluations.
However, the CPU can process operations significantly faster than the GPU, and as the number of sequential computations increases, the advantage of having more parallel threads is reduced.

At the scale of parallelization typical for RL applications (2,000 - 8,000 environments), the overhead from transferring data between host and device memory is the largest bottleneck. 
By keeping data entirely on the GPU, \cusadi\ enables speedups from 100-1000x in this regime.
For the applications in \autoref{sec:applications}, the estimated speedups are shown in \autoref{tab:speedup}.

\begin{table}[tb]
\caption{Estimated speedups compared to parallel CPU evaluation for the applications in \autoref{sec:applications}, based on the batch size used and function complexity.}
\begin{center}
\begin{tabular}{| c | c | c| c |} 
 \hline
 Application & Batch Size & N. instr. & Speedup [w/o data transfer]  \\ [0.5ex] 
 \hline\hline
 \autoref{sec:application_MPC} & 4096 & 1E5 & 14.38x [1.54x] \\ 
 \hline
 \autoref{sec:application_training} & 4096 & 1E4 & 104.84x [1.98x] \\
 \hline
 \autoref{sec:application_drone} & 10000 & 1E4 & 143.18x [2.98x] \\
 \hline
\end{tabular}
\end{center}
\label{tab:speedup}
\end{table}
\section{Applications}
\label{sec:applications}
We present several examples demonstrating how \cusadi\ can be used for robotics.
However, any application with repeated functional substructures at large scales (value iteration, fluid/weather simulation, image processing, finite element analysis, etc.) could leverage this framework for efficient GPU parallelization.

We consider two systems: the MIT Humanoid \cite{Saloutos2023_MITHumanoid} (\autoref{sec:application_MPC}, \autoref{sec:application_training}) and a planar, thrust-limited quadcopter (\autoref{sec:application_drone}).

\subsection{MPC Parallelization for the MIT Humanoid}
\label{sec:application_MPC}
A limitation of the SIMD parallelization approach described in \autoref{section:cusadi} is handling complex branching logic during function evaluation.
The vectorization is only valid when every function instance has the same set of instructions that can be performed in lock step, and branching evaluation paths can break this synchronized parallelization.
While simple ternary statements can be parallelized (e.g., \texttt{min} and \texttt{max} operators), functions with diverging evaluation paths are challenging to evaluate synchronously.
Consequently, \cusadi-parallelizable functions must have a finite set of synchronous instructions, limiting them to be closed-form and relatively free of branching logic.

Unfortunately, the algorithms to solve OCPs typically involve conditional divergence at each solver iteration, such as checking for solution convergence, line search criteria, and/or constraint violations.
However, prior work has shown that \textit{approximations} of an OCP are typically ``good enough" to achieve stable closed-loop performance for robotic systems.
The accuracy and convergence criteria of the OCP can be relaxed significantly without sacrificing controller quality, as in \cite{Diehl2005_realtimeIteration, Grandia2023_perceptiveLocomotion, Numerow2024_RobustSuboptimalMPC, Khazoom2024_TailoredMPC}.

Consequently, we approximate the solution to an OCP with a \textit{strictly fixed} number of operations.
Suppose a single solver iteration can be expressed in closed-form as 
\begin{equation}
    \label{eq:solver_iteration}
    \Xvec_{k+1} = h(\Xvec_k),
\end{equation}
where $\Xvec_k \in \Re^{N}$ is the current solution iterate of an OCP and $h: \Re^{N} \rightarrow \Re^{N}$ is a single iteration of some arbitrary solver.
Then we can express an approximate solution $\hat{\Xvec}$ to the optimization by recursively applying \eqref{eq:solver_iteration} $M$ times, so that $\hat{\Xvec} = H(\Xvec_0) :=  h^{M}(\Xvec_0)$.
This eliminates branching logic within the solver, allowing us to express $H$ as a \cusadi\ expression.
One advantage of this approach is that the accuracy of the solver can be tailored for computational demands as necessary.

With this approximation, we seek to solve the OCP in \autoref{sec:SQP} symbolically for parallelization.
Using the previous solution as an initial guess, a single QP iteration is often sufficient to approximate the solution (a ``real-time iteration", as in \cite{Diehl2005_realtimeIteration}).
This reduces \eqref{eq:OCP_general} to a QP problem.

\begin{figure}[t]
    \centering
    \includegraphics[width=0.95\linewidth]{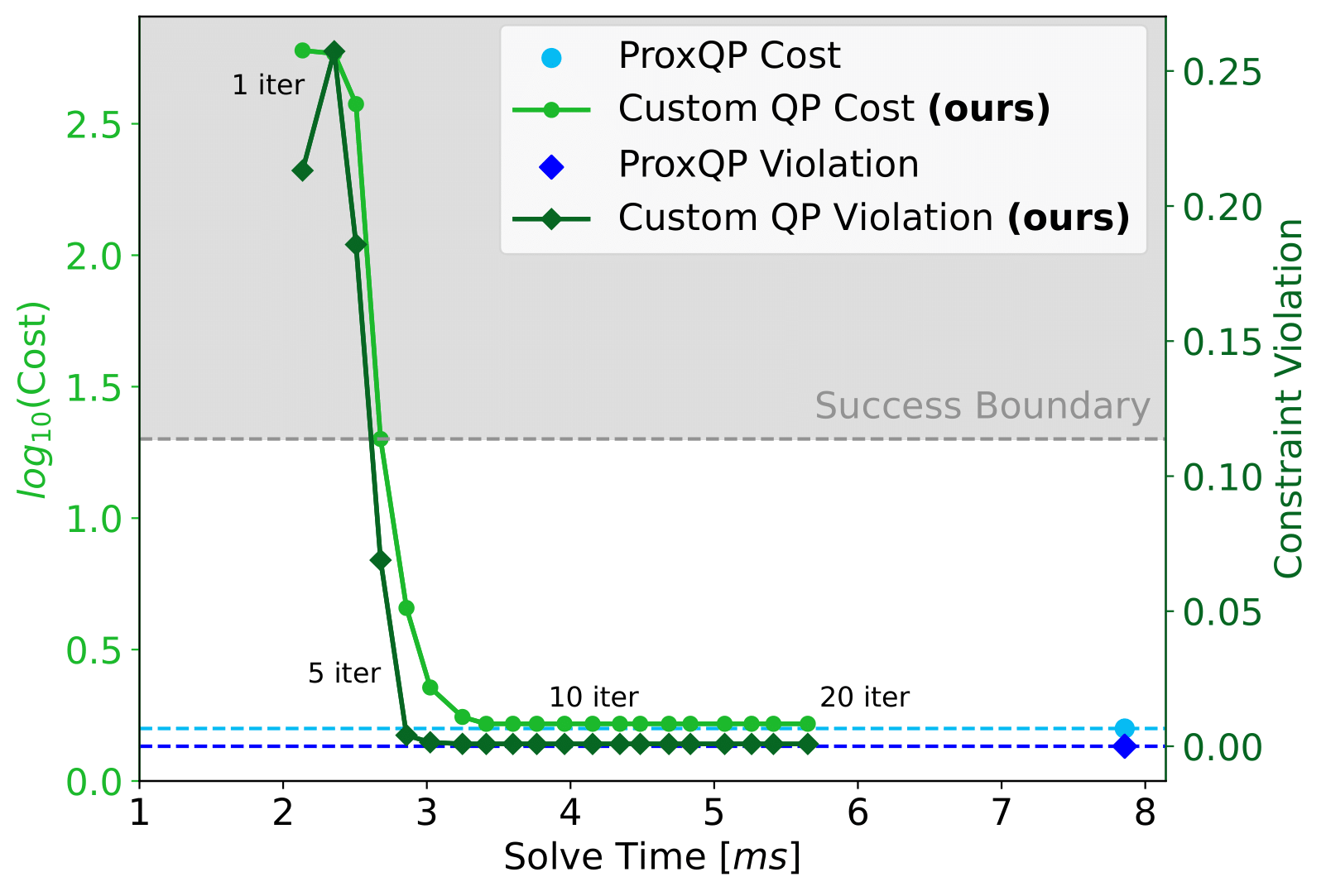}
    \caption{Pareto curve of closed-loop cost and constraint violation vs. evaluation time in closed-loop simulation for a single environment. The ``ground truth" for the QP is computed with \texttt{ProxQP} \cite{Bambade2023_ProxQP}, and the grey area represents when the controller is no longer closed-loop stable in IsaacGym.}
    \label{fig:pareto_MPC}
\end{figure}

To solve the QP, we adopt a \textit{penalty-based} method to approximate the original QP problem in~\eqref{eq:SQP_subproblem} by incorporating inequality constraints into the cost function, penalizing solution deviations from feasibility. The approximated QP problem can be represented as follows:
\begin{eqnarray}
\label{eq:cost_penalty}
    \min_{\delta\mathbf{z}} &&  \frac{1}{2} \delta\mathbf{z}^T \mathbf{P}_k \delta\mathbf{z}
                                + \mathbf{c}_k^T \delta\mathbf{z} + \mu_{k} \cdot {p}(\mathbf{A}_k^{ineq} \delta\mathbf{z} - \mathbf{b}_k^{ineq}) \nonumber\\
    \text{s.t.}             &&  \mathbf{A}_k^{eq} \delta\mathbf{z} = \mathbf{b}_k^{eq},
\end{eqnarray}
where $\mu$ is a penalty parameter, and $p(\cdot): \Re^{M_{ineq}} \rightarrow \Re$ is a penalty function, such as a quadratic function or an $l_{1}$ penalty function. It can be shown that for sufficiently large $\mu$, the solutions of the approximated problem also solve the original problem~\cite{Boyd2004_ConvexOptimization}. By iteratively increasing the penalty parameter to a sufficiently large value (e.g., $\mu_{k+1}= \alpha \mu_{k}, \alpha > 0$), the solution gradually converges to the original problem. 
The equality-constrained problem in~\eqref{eq:cost_penalty} can be solved by applying the \textit{LDL$^{T}$} factorization approach \cite{Boyd2004_ConvexOptimization} to the KKT equations. 
Therefore, we chose the formulation in~\eqref{eq:cost_penalty}, as it allows us to obtain accurate equality-constrained solutions with minimal computational overhead.

For the MIT Humanoid, we implement the single rigid-body model (SRBM) nonlinear MPC controller detailed in \cite{Hong_MPCSO3} entirely in \casadi, and demonstrate its subsequent \cusadi\ parallelization across 4,096 environments in IsaacGym, as shown in \autoref{fig:parallel_MPC}.
For the penalty function, we use ${p}(\mathbf{x}) = \sum_{i=1}^{M_{ineq}} (\max(0, x_{i}))^2$.

There is a direct trade-off between the convergence accuracy of the approximated MPC and the evaluation time of the function.
With too few, the controller fails to be stable in closed-loop simulation, but past a certain number of iterations, the marginal benefit of each solve diminishes rapidly while incurring significant computational cost, as visualized in \autoref{fig:pareto_MPC}.
The fidelity of the controller can be tuned to accommodate the computational demands of the application, such as sampling high-quality rollouts offline or embedding MPC in RL training.

Jenelten \textit{et al.}\cite{Jenelten2024_DTC} requires roughly 14 seconds per PPO iterations when trained with 4,096 environments at a 200 Hz simulation frequency, 50 Hz policy frequency, and 2.2 Hz MPC frequency.
While we leave learning a policy alongside the parallelized MPC to future work, initial tests in IsaacGym showed an iteration time of roughly 1.24 seconds per PPO iteration with the same frequencies, corresponding to a speedup of roughly 11x.
While the MPC controller in \cite{Jenelten2024_DTC} is more complex than our SRBM MPC, leveraging the GPU and eliminating the overhead of data transfer significantly improves the efficiency of learning with optimization in the loop.

\subsection{RL Training with Centroidal Momentum}
\label{sec:application_training}

\begin{figure}[tb]
    \centering
    \includegraphics[width=1.0\linewidth]{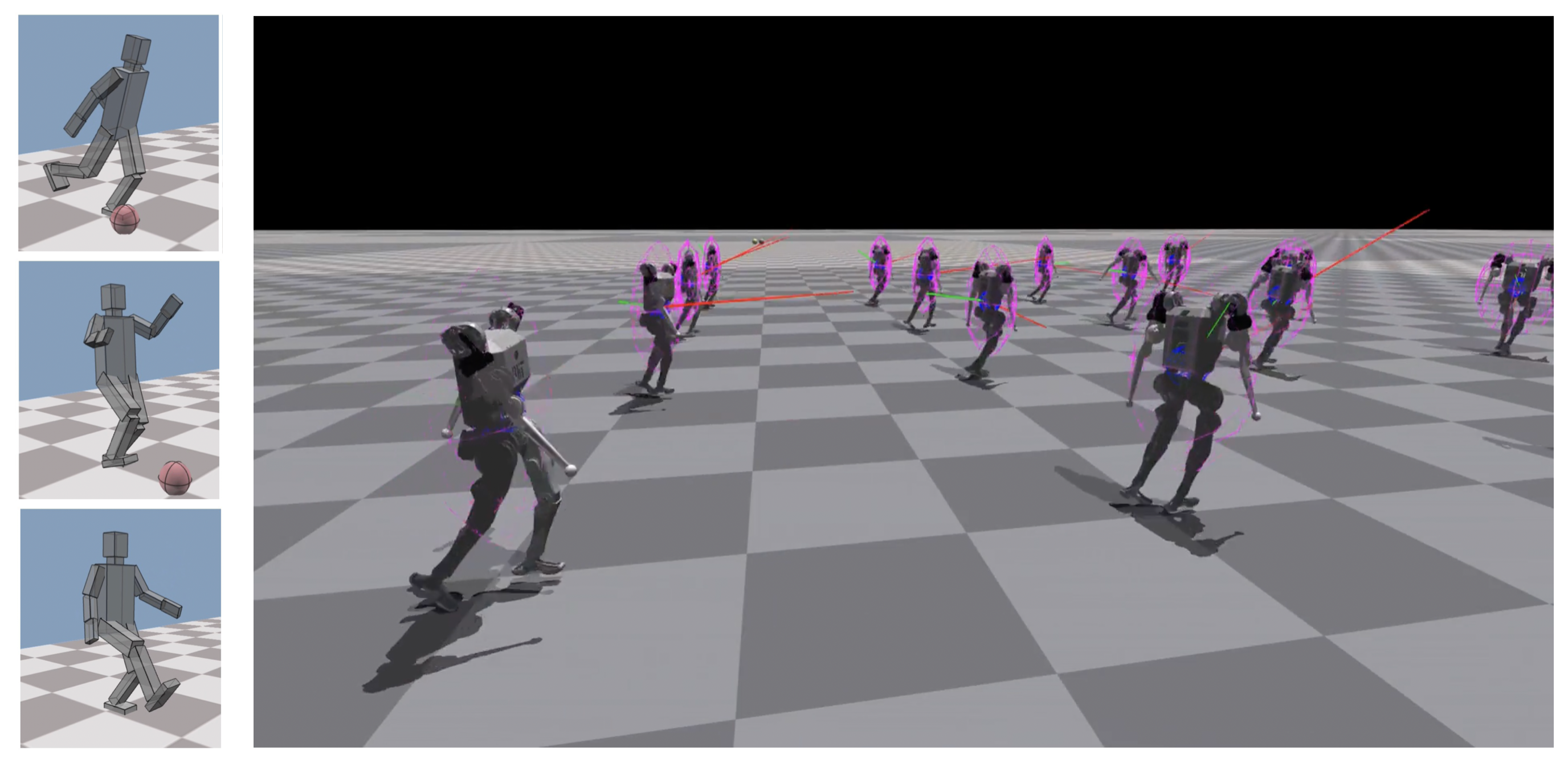}
    \caption{\textit{Left:} An example of how tracking centroidal angular momentum can generate natural behavior from the legs and arms for a humanoid robot \cite{Wensing2016_centroidal}. \textit{Right:} Using \cusadi, we rewarded tracking a desired centroidal angular momentum based on \cite{Wensing2016_centroidal}, instead of a desired base angular velocity. Emergent arm swing is noticeable. We also use \cusadi\ to visualize the centers of mass (blue), composite inertia (pink), angular momentum (green) and linear momentum (red).}
    \label{fig:arm_swing}
\end{figure}

We demonstrate how \cusadi\ can act as a bridge to incorporate in model-based quantities relevant for legged locomotion (centroidal momentum \cite{Orin2013_centroidal}, center of pressure \cite{Sentis2009_COP}, the divergent component of motion \cite{Englsberger2015_DCM}, composite rigid-body inertia, etc.) to RL settings.
While these could be computed directly in the RL environments, it can be challenging to efficiently implement the necessary algorithms across tensors of state data, especially if sparsity can be exploited.
These quantities only need to be expressed symbolically for a \textit{single} instance (made straightforward with the dynamics libraries mentioned in \autoref{section:casadi}) to be computed in parallel across any number of environments.

Taking inspiration from \textcite{Wensing2016_centroidal}, we parallelize computing the centroidal momentum matrix (CMM) for the MIT Humanoid (using the \casadi-compatible dynamics algorithms in \texttt{spatial\_v2}) to augment RL training, as shown in \autoref{fig:arm_swing}.
Typically, tracking some desired angular velocity for the base is rewarded in RL settings.
For this simple example, we instead reward tracking a desired centroidal angular momentum.
By doing so, we observe emergent arm swing during locomotion, corroborating the relationship between minimizing the CAM and arm motion from the original work, as well as \cite{Khazoom2022_armSwing}.

\subsection{Parallelized Rollouts}
\label{sec:application_drone}

\begin{figure}[tb]
    \centering
    \includegraphics[width=1.0\linewidth]{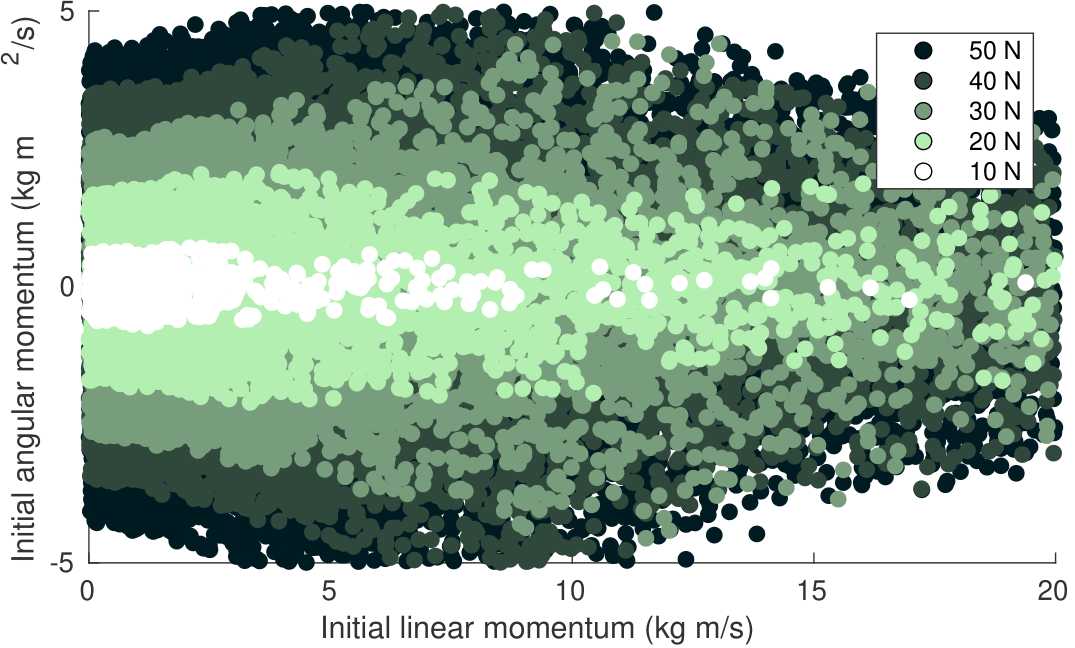}
    \caption{Region of attraction from Monte Carlo simulations of quadcopter with LQR controller, across various thrust limits.}
    \label{fig:drone_ROA}
\end{figure}

\begin{figure*}[tb]
    \centering
    \includegraphics[width=1\textwidth]{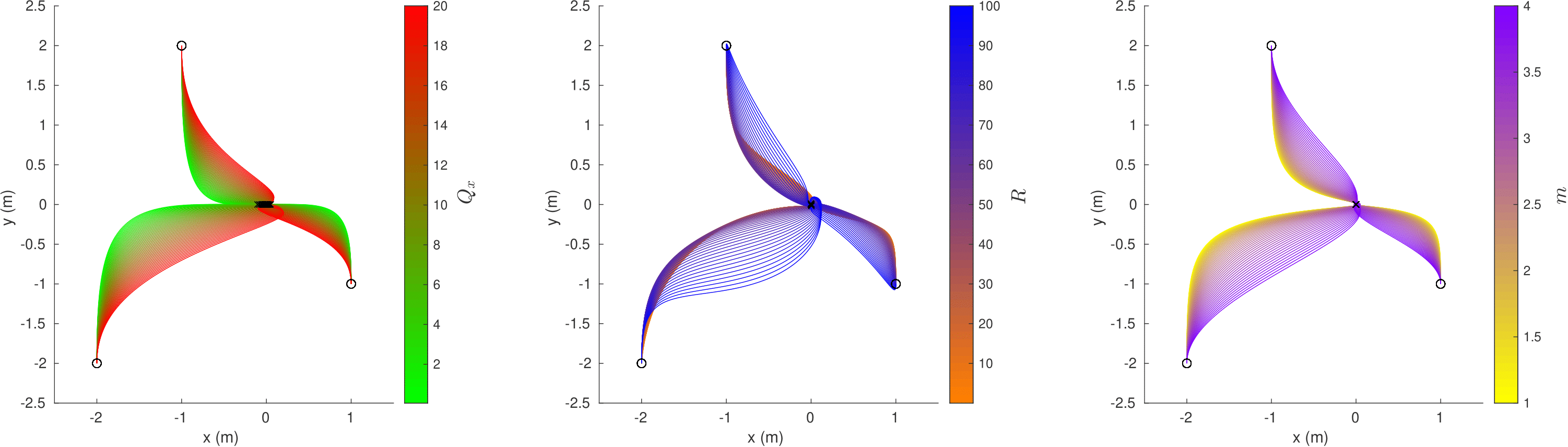}
    \caption{Sweep of controller and inertial parameters on the LQR trajectory.The resolution of the sweep allows us to see finer details of the trajectory and when they occur (e.g. the overshooting in $x$ from a large $Q_x$). These sweeps could be evaluated in parallel online to quickly sample and avoid potentially undesirable states.}
    \label{fig:drone_sweep}
\end{figure*}

The system we consider is a planar quadcopter subject to thrust limits.
We consider two scenarios to showcase the parallelization of \cusadi.
First, given a controller, from what initial states can the system be stabilized?
What is the region of attraction of that controller?
Second, given an initial and desired state, how is the optimal trajectory affected by the system and controller parameters?
Can they be adjusted to meet design or state constraints for the system?

With \cusadi, we parallelize a closed-loop simulation step of the quadcopter with a linear quadratic regulator (LQR) controller.
If the LQR horizon $T$ is finite, the problem can be rewritten as an equality-constrained quadratic program, and its KKT system can be solved with symbolic \textit{LDL$^{T}$} decomposition as in \autoref{sec:application_MPC}.
For the infinite-horizon case, the structured doubling algorithm from \textcite{Wang2008_fastApproxMPC} can be implemented to solve the discrete algebraic Riccati equation (DARE), which has quadratic convergence to the solution $S_\infty$.
While the algorithm should be repeated until convergence, we can again fix the number of iterations to approximate the solution.
For the drone, we found that 10-15 iterations of this algorithm was suitable.

The symbolic LQR solution is used as the input for the drone dynamics, and integrated with the semi-implicit Euler scheme in \casadi.
In addition to the current state, we specify the thrust limits, inertial parameters, and LQR weights as additional parameters for the function.
Overall, our \casadi\ function computing the closed-loop dynamics with the LQR controller takes the form
\begin{equation}
    \mathbf{z}_{k+1} = f_{quad}(\mathbf{z}_k, \theta),
\end{equation}
with quadcopter state $\mathbf{z} \in \Re^{6}$ and parameters $\theta$.

With \cusadi, we parallelize this closed-loop simulation step for the quadcopter and rollout 10,000 environments in parallel.
For the first scenario, we fix the controller and initialize each environment with different angular and linear momenta with zero position and rotation offset, and determine which rollouts were stable.
The resultant region of attraction for each thrust limit is visualized in \autoref{fig:drone_ROA}.

The MPC described in \autoref{sec:application_MPC} could likewise be rolled out across large batches to estimate regions of stability in state space for the humanoid, a high-dimensional problem that would be extremely inefficient to compute without GPU parallelization.

For the second scenario, the quadcopter is initialized from rest at some non-zero position, and the parameters of the system are varied to study their effects on the optimal trajectory, as shown in \autoref{fig:drone_sweep}.
The resultant rollouts directly visualize the effect of the parameters on closed-loop simulation, making them much easier to tune.
A potential use case for \cusadi\ is performing these kinds of sweeps \textit{online}.
Similar to \cite{Sacks2022_LearningMPPI}, Monte Carlo rollouts can be evaluated on the GPU to better estimate uncertain parameters of the system or adjust its trajectory in real time.

While these low-dimensional examples do not require GPU scaling for evaluation, they serve to illustrate how \cusadi\ can be used to tackle high-dimensional problems that require substantial data.

\section{Conclusion}
In this work, we extend the symbolic framework of \casadi\ so that arbitrary closed-form expressions can be parallelized on the GPU with \texttt{CUDA} and formulate a closed-form approximation to the OCP to evaluate MPC in parallel at a large-scale.

As a tool, \cusadi\ can be extended in several ways.
Parallelism \textit{within} individual expressions could also be exploited, especially for larger problems as studied in \cite{Plancher2019_DDPGPU}.
Results from graph theory could be used to identify parallelization opportunities from \casadi\ expression graphs.
However, this would have to be balanced against the overhead of starting and synchronizing additional threads.


For future work, the parallelization offered by \cusadi\ opens up several promising directions.
To improve the locomotion capabilities of the MIT Humanoid, we plan to learn a \textit{residual policy} alongside the parallelized MPC with reinforcement learning \cite{Silver2018_ResidualPolicy}. 
Another potential direction is to learn the value function for MPC with parallelized rollouts.
The function could then be used to bootstrap value estimates in RL pipelines, similar to \cite{Grandesso2023_CACTO}, or as a terminal cost for more complex MPC controllers.



\addtolength{\textheight}{0cm}
\printbibliography
\end{document}